\documentclass[utf8]{frontiersSCNS} % for Science, Engineering and Humanities and Social Sciences articles
\usepackage{url,hyperref,lineno,microtype,subcaption}
\usepackage[onehalfspacing]{setspace}
\usepackage{booktabs}
\usepackage{mathptmx}

\usepackage{multirow}
\usepackage{xcolor}
\usepackage[normalem]{ulem}

\def\keyFont{\fontsize{8}{11}\helveticabold }
\def\firstAuthorLast{Pan {et~al.}} %use et al only if is more than 1 author
\def\Authors{Hoang Long Nguyen\,$^{1,*}$, Zhenhe Pan\,$^{1,*}$, Hashim Abu-gellban$^{1}$, Fang Jin$^{2}$, and Yuanlin Zhang$^{1}$}

\def\Address{
$^{1}$Department of Computer Science, Texas Tech University \\
$^{2}$Department of Statistics, George Washington University 
}
\def\corrAuthor{Zhenhe Pan}

\def\corrEmail{zhenpan@ttu.edu}

\begin{document}
\onecolumn
\firstpage{1}

% \title[Google Trend Analysis of Covid'19]{\theName: Spatio-Temporal Google Trend Analysis of Coronavirus Pandemic (Covid'19)} 

\title[Google Trends Analysis of COVID-19]{Google Trends Analysis of COVID-19 Pandemic} 

\author[\firstAuthorLast ]{\Authors} %This field will be automatically populated
\address{} %This field will be automatically populated
\correspondance{} %This field will be automatically populated

% \extraAuth{}% If there are more than 1 corresponding author, comment this line and uncomment the next one.
\extraAuth{Hoang Long Nguyen \\ long.nguyen@ttu.edu}

\maketitle

\begin{abstract}

% The first suspected case of the novel coronavirus disease was discovered on December 1, 2019. Later, the 
The World Health Organization (WHO) announced that COVID-19 was a pandemic disease on the 11th of March as there were 118K cases in several countries and territories. Numerous researchers worked on forecasting the number of confirmed cases
% , since COVID-19 had gravely impacted the health and economics. Anticipating 
since anticipating the growth of the cases helps governments adopting knotty decisions to ease the lockdowns orders for their countries. These orders help several people who have lost their jobs and support gravely impacted businesses. 
Our research aims to investigate the relation between Google search trends and the spreading of the novel coronavirus (COVID-19) over countries worldwide, to predict the number of cases.
We perform a correlation analysis on the keywords of the related Google search trends according to the number of confirmed cases reported by the WHO. After that, we applied several machine learning techniques (Multiple Linear Regression, Non-negative Integer Regression, Deep Neural Network), to forecast the number of confirmed cases globally based on historical data as well as the hybrid data (Google search trends). Our results show that Google search trends are highly associated with the number of reported confirmed cases, where the Deep Learning approach outperforms other forecasting techniques. We believe that it is not only a promising approach for forecasting the confirmed cases of COVID-19, but also for similar forecasting problems that are associated with the related Google trends.

% With the increment of social networking and the demand for search over internet, the information retrieval has produced another aspect of data source for analysis, so called search trends. In addition, it is more interesting if the analysis provides patterns and trends based on geographical and temporal occurrence, and also keep track of how they change over time. Thus, the aim of this study is to investigate the relation of Google search trends and the spreading of the novel coronavirus (Covid'19) over countries worldwide. The keywords employed for Google Trends are collected from user definitions and extended by related words recommended by the search trends. The data associated with these keywords are used for correlation analysis with the confirmed cases reported by the World Health Organization (WHO). Additionally, several machine learning techniques are also employed to forecast the number of confirmed cases based of historical data and the hybrid data which has the addition of Google search trends. Our results show that Google search trends are highly associated with the number of reported confirmed cases. We believe that it is not only a promising alternative for Covid'19 confirmed cases forecast, but also many other forecasting scenarios. 

\tiny
 \keyFont{ \section{Keywords:} coronavirus, covid'19, forecasting, search trends, neural networks, machine learning, spatio-temporal analysis} %All article types: you may provide up to 8 keywords; at least 5 are mandatory.
\end{abstract}

\section{Introduction}
\begin{figure}
    \centering
    \includegraphics[width=0.85\linewidth]{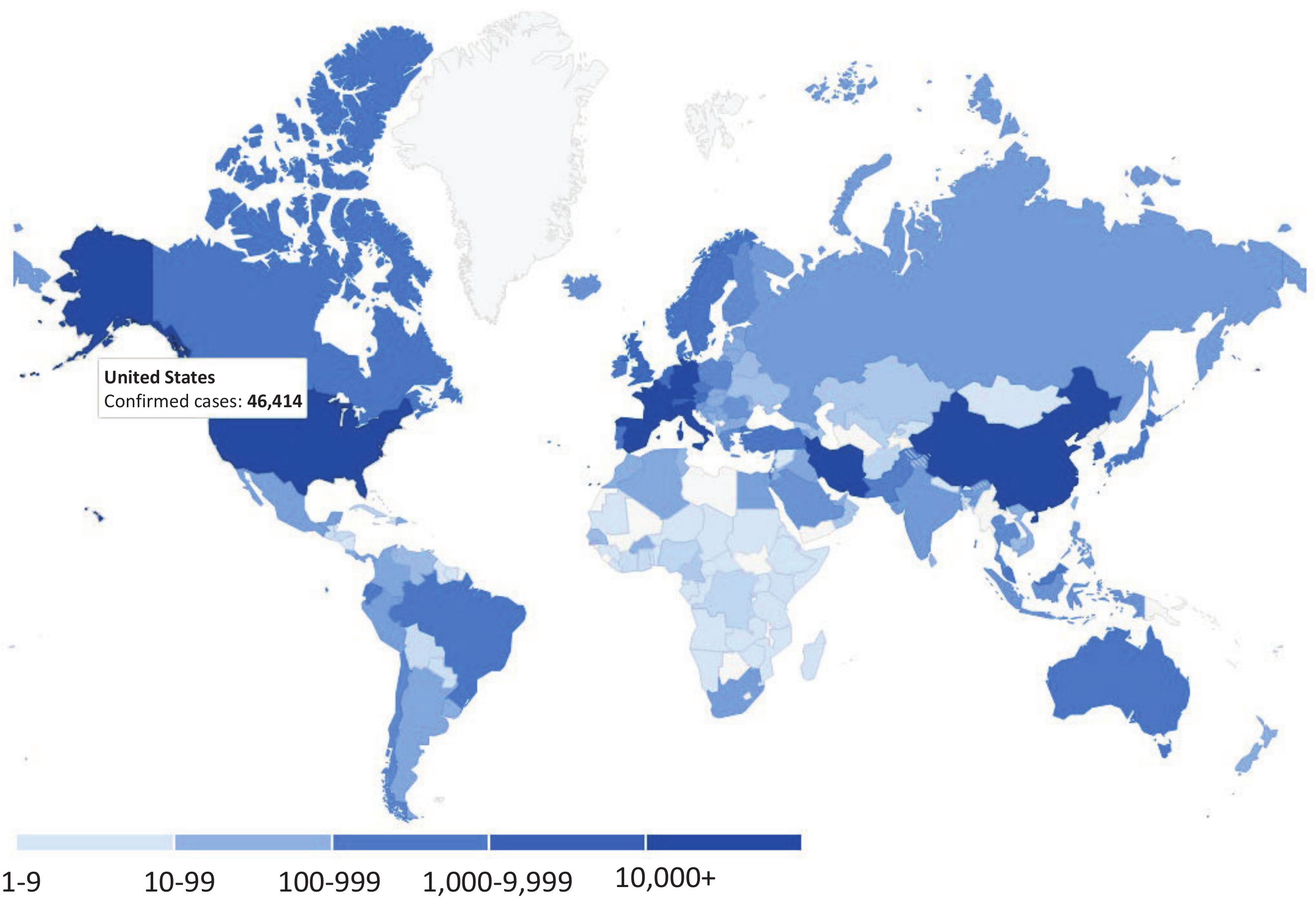}
    \caption{Coronavirus Confirmed Cases Worldwide \cite{googlecovid}}
    \label{fig:global-coronavirus}
\end{figure}

Since the outbreak of coronavirus in December, 2019 in Wuhan, China, COVID-19 is spreading exponentially and has already effected nearly every county in the world, infecting millions of people and causing more than tens of thousand deaths around the world (as of March 16, 2020), as shown in Figure \ref{fig:global-coronavirus}. It has caused extremely catastrophic social and economic damage throughout the world. Coronavirus job losses could total $47$ million, the unemployment rate may hit $32\%$,  according to a Federal Reserve estimate.\footnote{https://www.cnbc.com/2020/03/30/coronavirus-job-losses-could-total-47-million-unemployment-rate-of-32percent-fed-says.html} To predict the infected patient number is crucially important to both individual and decision makers  preparedness, and to flatten the curves. However, how to accurately predict the number of infected patients is never a trivial task. There are numerous factors contribute to this virus's propagation, such as population mobility, temperature, and medical condition. 

Ferguson et al.~\cite{ferguson2020report} applied a previously published microsimulation model to the UK and the US dataset, and concluded that to flatten the curve requires a combination of social distancing of the entire population, home isolation of cases and household quarantine of their family members. The authors also estimated that up to $2.2$ million people could die if no actions were taken to stop transmission in the US. Using another statistical model, Murray et al.~\cite{covid2020forecasting} predict that the US infected patient number would peak around April 15. At this peak date, the US is projected to need $220,643$ total hospital beds ($32,976$ for ICU), and $26,381$ ventilators to support COVID-19 patients. Nationwide COVID-19 deaths are predicted to also peak on April 15, escalating to $2,214$ deaths per day on average. Nationwide, the mean value of the total COVID-19 deaths is projected at about $84,000$. 

Unfortunately, most of the existing model based prediction approaches rely on some oversimplified assumptions such as virus travel distance, and timely and effective quarantine measures. However, these assumptions are rarely justified because the social structure is widespread. In addition, since this virus is still novel to the human being, and there are still so many unknown about spreading patterns, severity, and many more, which may introduce high irreducible error~\cite{james2013introduction}. For example, Lydia Bourouiba~\cite{ferguson2020report} recently demonstrated a Respiratory Emissions model is much more complicated than the traditional established model, and the peak exhalation speeds can reach up to $33$ to $100$ feet per second, creating a cloud that can span approximately $23$ to $27$ feet, which is far larger than the current recommended social distancing (around $6$ feet). A $2020$ report from China demonstrated that severe acute respiratory syndrome coronavirus 2 (SARS-CoV-2) virus particles could be found in the ventilation systems in hospital rooms of patients suggesting these virus particles can travel long distances from patients.% There is a need to understand the biophysics of host-to-host respiratory disease transmission accounting for in-host physiology, pathogenesis, and epidemiological spread of disease. The rapid spread of COVID-19 highlights the need to better understand the dynamics of respiratory disease transmission by better characterizing transmission routes, the role of patient physiology in shaping them, and best approaches for source control to potentially improve protection of front-line workers and prevent disease from spreading to the most vulnerable members of the population.

Nowadays, more and more people have access to the internet, and to search for information that are closely related to their daily lives, feeling, and thoughts. It is estimated that there are around $63,000$ Google searches per second. The average person makes some three or four searches every day.\footnote{https://serpwatch.io/blog/how-many-google-searches-per-day/} 
% on the social media platforms, such as Facebook, tweeter, and etc. For example, there are 2.2 billion users access Facebook, Instagram, WhatsApp or Messenger every day, and 2.8 billion use one of this family of apps each month~\footnote{https://zephoria.com/top-15-valuable-facebook-statistics/}. These social media users generate a huge amount of data, making internet a revealing data source to study our real world, to forecast and to make prediction about the future. Thus, the combination of historical time series data with specific social media data can generate better prediction performance than using historical data alone. A wide range of applications resort to social media for a better predictive models. 
Google Trends is a website sponsored by Google that analyzes the popularity of top search queries in Google Search across various regions and languages. Since more and more people are having access to the internet, and are more reliable than ever to search for information that they care about. Thus, Google trends are revealing and can provide an opportunity to examine people's concerns as well as hot topics that they are interested with. Researchers have used Google Trends data to investigate a number of researches such as: (1) disease outbreak prediction. As far as we know, Carneiro and Mylonakis~\cite{carneiro2009google} are the first authors to introduce the more generic Google Trends tool to health professionals, to show how they can track disease activity. Verma et al.~\cite{verma2018google} illustrated there is a strong temporal correlation between some diseases (chikungunya, dengue fever, and Haryana), and Google search trends.  Zhang et al.~\cite{zhang2018using} used Google Trends and ambient temperature to predict seasonal influenza outbreaks, and suggested internet search metrics in conjunction with temperature can be used to predict influenza outbreaks. (2) economy and financial market prediction. Hong et al.~\cite{pai2018using} used Internet Search Trends and Historical Trading Data for Predicting Stock Markets, and showed that using hybrid data can provide more accurate forecasting results than using single historical trading data. MY Huang et al.~\cite{huang2019forecasting} found that the utilization of Google search data allows us to construct a model to forecast directional movements in the S\&P 500 index. Preis et al.~\cite{preis2013quantifying} suggested that Google Trends data does not only reflect aspects of the current state of the economy, but may have also provided some insight into future trends in the behavior of economic actors with other concepts in technical analysis. (3) political election. A group of researchers at Wellesley College examined data from Google Trends data successfully predicted the outcome in 33.3\% of cases in 2008 and 39\% in 2010. By analyzing data from Google Trends, \cite{mavragani2016yes} calculated a valid approximation of the final result, thus contributing to the discussion of using Google Trends as an elections' results prediction tool in the future.

In this paper, we explore the Google trends data to derive its relationships with the COVID-19 spreading situations. Instead of focusing on model based prediction, we propose to use Google Trends data and combine with the historical time series for future cases prediction. Our approach is pure data driven, and skips the complicated mathematical modeling, which greatly reduces the algorithm complexity. We did comprehensive experiments, and applied multiple popular prediction models, such as multiple linear regression model, statistical model, and deep neural network on worldwide data to see the correlation between search trends and infected cases. Our experiments demonstrated that there is a strong relationship between infected patient cases and Google trends data, and can be used with other analysis techniques for a better understanding of this mysterious disease spreading. The contributions of this paper can be summaries as followings:
\begin{itemize}
  \item To the best of our knowledge, we are the first to use Google trends data to predict the number of confirmed coronavirus cases utilizing different model types: Linear model, Statistical model, and Deep learning model. 
  \item We present performance comparison across three models either using Google trends or not using Google trends feature. The results show that Google trends play an important role in the performance of the prediction models.
\end{itemize}

\section{Related Work}
There is a large number of studies about using Google Trends in forecasting algorithms. Here, we discuss two themes: The studies of Google Trends for disease control analysis and for other application domains.
\vspace{6mm}
\subsection{Google Trends for disease control analysis}
\cite{cook2011assessing} assessed Google Flu Trends\footnote{https://www.google.org/flutrends} performance in the United States during the 2009 influenza virus A (H1N1) pandemic. The assessment showed that the internet search behavior changed during pH1N1. And the updated version of the Google Flue Trends technique performed better than the prior one \cite{yang2015accurate}.
\cite{anggraeni2016using} used Google Trends data to build a forecasting model by applying the Autoregressive Integrated Moving Average with exogenous variable (ARIMAX) method, to predict the number of dengue fever cases in Indonesia. \cite{xu2017forecasting} used Google search queries to build a statistical model to anticipate the number of influenza cases in Hong Kong. They compared different forecasting approaches: (Generalized Linear Model (GLM), Least absolute shrinkage and selection operator (LASSO), ARIMA, Feed Forward Neural Networks (FNN), and Bayesian model averaging (BMA). Authors recommended using FNN to predict the cases with better accuracy. Similarly, \cite{ginsberg2009detecting} used search engine queries to estimate weekly influenza activity in each region of the United States with a reporting lag of about one day compared with one to two weeks in traditional surveillance system. \cite{silva2019googling} proposed a hybrid neural network approach named "Denoised NNAR" combining Neural Network AutoRegression (NNAR) with the singular spectrum analysis. The analysis used Google Trends to reduce noise in fashion data. \cite{shaghaghi2020influenza} built a model using Long Short-Term Memory (LSTM) to anticipate the number influenza cases using the data of flu season from Centers for Disease Control and Prevention (CDC) and World Health Organization (WHO), and Google Trends to help the decision maker increasing or decreasing vaccines and medicines in advance. \cite{kapitany2019can} stated that Google Trends were very correlated with the Lyme disease incidence report in Germany. 
% Our work is distinct because COVID-19 has become a pandemic. It has been spreading dramatically and gotten attention worldwide. 
\vspace{5mm}
\subsection{Google Trends for other application domains}
\cite{bokelmann2019spurious} confirmed that there was an increment of work using Google Trends in tourism research. In this work, Google Trends was used as a predictor for short-term tourism demand. There were several traditional forecasting techniques which were utilized demonstrated that Google Trends played a significant role in short-term forecasting of tourism demand. Similar tourism related topics were discussed by \cite{bangwayo2015can,park2017short,onder2017forecasting}. \cite{xu2014stock} showed that combining the time series analysis algorithms with Google Trends and Yahoo finance improved forecasting the the stock prices. \cite{askitas2009google} and ~\cite{bulut2018google} stated that Google Trends were useful in predicting numerous economic variables (e.g., unemployment, exchange rates). \cite{yu2019online} worked in predicting Ford car sales in Argentina using GT. They used the keyword “Ford”, to improve their forecasting model. Even though there were some traps happened in the past of using Google Trends in Big Data analysis that discussed by \cite{lazer2014parable} and \cite{butler2013google}, however, many improvements have been done by Google since then. Additionally, the studies of Google Trends analysis for disease control are still increasing through time.
\vspace{-3mm}
\section{Method}
% \subsection{Preliminary of Regression Analysis}
% Regression analysis is a widely used statistical technique to find relationships between independent and dependent variables~\cite{seber2012linear}. It searches for a set of parameters for a model to minimize the total errors on the training dataset. Regression analysis models the output $Y$ is a function, $f$, of a set of input variables $x$, which can be written as 
% \begin{equation}
% Y = f(x, \beta) + \epsilon
% \end{equation}, where $\beta$ is the function parameters, and $\epsilon$ is the unobserved determinants of $Y$ or random statistical noise. Typical function of $f$ includes linear function and logistic function. The training process is to find the optimal function parameters, $\beta$, which minimizes the total error across the training instance. 

% \vspace{8mm}
\subsection{Building Feature for Regression Models}
We collect 13 different Google trends features based on the 13 search keywords. Some of these features might be revealing, and others might contain strong noise which is not suitable for future prediction. We differentiate those features into two classes base on the Pearson method \cite{benesty2009pearson}. This approach has shown its success in some similar work as done by \cite{nguyen2019self}. The Pearson correlation coefficient attempts to measure the similarity (correlation coefficient) between a series and the original one. Given two series $X$ and $Y$, it can be defined as:
\begin{equation}
    corr(X, Y) = \frac{ \sum_{i=1}^{T}{(x_i-\overline{x})*(y_i-\overline{y})}}{\sqrt{\sum_{i=1}^{T}{(x_i-\overline{x})^2}}*\sqrt{\sum_{i=1}^{T}{(y_i - \overline{y})^2}}},
\end{equation}
where $\overline{x}$ and $\overline{y}$ is the mean value of the time series $X$ and $Y$ respectively. 
The correlation coefficient values which are greater than $0.7$ are treated as highly correlated and used as features for the prediction model; while the values which are below that threshold are considered as noise and being ignored.
% \vspace{-17mm}
\begin{table*}[htbp]
  \centering
 \caption{Correlation between Google trends of search queries and confirmed cases worldwide.}
    \begin{tabular}{lc|c|rr|r}
    \toprule
          & \textbf{Correlation} & \textbf{p-value} &       & \multicolumn{1}{c|}{\textbf{Correlation}} & \multicolumn{1}{c}{\textbf{p-value}} \\
    \midrule
    cases of covid19 & 0.8633 & 7.23E-19 & \multicolumn{1}{l}{coronavirus update} & \multicolumn{1}{c|}{0.7796} & \multicolumn{1}{c}{2.17E-13} \\
    corona & 0.7789 & 2.34E-13 & \multicolumn{1}{l}{covid} & \multicolumn{1}{c|}{0.8650} & \multicolumn{1}{c}{5.14E-19} \\
    coronavirus & 0.7408 & 1.32E-11 & \multicolumn{1}{l}{covid 19} & \multicolumn{1}{c|}{0.8627} & \multicolumn{1}{c}{8.12E-19} \\
    coronavirus cases & 0.8196 & 1.18E-15 & \multicolumn{1}{l}{covid 19 cases} & \multicolumn{1}{c|}{0.8687} & \multicolumn{1}{c}{2.41E-19} \\
    coronavirus covid19 & 0.8174 & 1.62E-15 & \multicolumn{1}{l}{covid19} & \multicolumn{1}{c|}{0.8506} & \multicolumn{1}{c}{7.91E-18} \\
    coronavirus news & 0.7750 & 3.65E-13 & \multicolumn{1}{l}{covid19 cases} & \multicolumn{1}{c|}{0.8584} & \multicolumn{1}{c}{1.87E-18} \\
    \sout{coronavirus symptoms} & \sout{0.6664} & \sout{6.18E-09} &       &       &  \\
    \bottomrule
    \end{tabular}%
  \label{tab:correlation-trends-confirmed-cases}%
\end{table*}%
Table \ref{tab:correlation-trends-confirmed-cases} presents the correlation of Google trends using selected keywords with respect to the changes of new confirmed coronavirus cases. We can see that most of the selected keywords are highly correlated. The only keyword \textit{coronavirus symptoms} show less correlation coefficient. Therefore, we decided to drop Google trends by this keyword from the prediction models.

\vspace{2mm}
\subsection{Regression Model}
We study typical regression models from traditional approaches like the linear model, and statistical model such as negative binomial, to the most recent approach which is the deep neural network model.
\vspace{3mm}
\subsubsection{Multiple Linear Regression Model}
The most straightforward prediction model is the multiple linear regression model. Multiple linear regression attempts to model the relationship between two or more explanatory variables and a response variable by fitting a linear equation to observed data. Essentially, it can be considered as an extension of ordinary least-squares regression that involves more than one explanatory variable. Suppose there are $p$ distinct dependent variables, then the multiple linear regression model can be expressed as
\begin{equation}
Y=\beta_0+\beta_1X_{1}+\beta_2X_{2}+...+\beta_pX_{p}+\epsilon
\end{equation}, where for $X_i$ is the $i_{th}$ variable, and $\beta_i$ measures the association between $X_i$ and the response $Y$. Similarly with the linear regression model, the parameters, $\beta_0, \beta_1, ..., \beta_p$ here are the optimal estimators to minimize the sum of squared residuals, RSS. The multiple regression model is based on the following assumptions. 1) There is a linear relationship between the dependent variables and the independent variables. 2) The independent variables are not too highly correlated with each other. 3) $y_i$ observations are selected independently and randomly from the population. 4) Residuals should be normally distributed with a mean of 0 and variance $\sigma$, which is estimated as
\begin{equation}
\sigma^2=\frac{ \sum_{i=1}^{n}{e_i}^2}{n-p-i}
\end{equation}, where $e_i=y_i - \hat{y_i}$ is the residuals.

% \vspace{-2mm}
\subsubsection{Non-negative Integer Regression Model}
%\textcolor{blue}{Negative Binomial Regression}
Negative binomial regression is similar to regular multiple regression except that the dependent, $Y$ variable is an observed count that follows the negative binomial distribution. Negative binomial regression is a generalization of Poisson regression which loosens the restrictive assumption that the variance is equal to the mean made by the Poisson model. The traditional negative binomial regression model is based on the Poisson-gamma mixture distribution. This formulation is popular because it allows the modeling of Poisson heterogeneity using a gamma distribution. % The Poisson distribution may be generalized by including a gamma noise variable which has a mean of 1 and a scale parameter of $v$. The Poisson-gamma mixture (negative binomial) distribution that results is
Hilbe~\cite{hilbe2011negative} introduces the negative binomial distribution as:
\begin{equation}
\label{eqaution:bnd_definition}
\begin{aligned} p(y)&=P\left( Y=y|{u},\alpha\right) \\
&=\frac{\varGamma \left( y+\alpha^{-1} \right) }{\varGamma \left( y+\alpha^{-1} \right) }\left( \frac{ \alpha^{-1}}{\alpha^{-1}+\mu}\right) ^{\alpha^{-1}}\left( \frac{ \mu}{\alpha^{-1}+\mu}\right) ^y \end{aligned}
\end{equation}, where $\mu$ is the mean incidence rate of $Y$ per unit of exposure, and $\alpha$ is the heterogeneity parameter.
The traditional negative binomial regression model, designated the NB2 model in~\cite{hilbe2011negative}, is
\begin{equation}
\ln\mu=\beta_0+\beta_1x_{1}+\beta_2x_{2}+...+\beta_px_{p}
\end{equation}, where the predictor variables $x_1,x_2,...,x_p$ are given, and the population regression coefficients $\beta_0, \beta_1, \beta_2,..., \beta_p$ are to be estimated. Given a random sample of $n$ subjects, for observed subject $i$, the dependent variable is $y_i$, and the predictor variables are $x_{1i},x_{2i},...,x_{pi}$. We denote $x_i=(x_{1i},x_{2i},...,x_{pi})$ and $\beta=(\beta_0, \beta_1, \beta_2,..., \beta_p)^T$, therefore, the Equ.~\ref{eqaution:bnd_definition} for an observation $i$ can be re-written as:
\begin{equation}
\begin{aligned} P\left( Y=y_{i}|{u}_{i},\alpha\right) =\frac{\varGamma \left( y_{i}+\alpha^{-1} \right) }{\varGamma \left( y_{i}+\alpha^{-1} \right) }\left( \frac{1}{1+\alpha\mu_i}\right) ^{\alpha^{-1}}\left( \frac{ \alpha\mu_{i}}{1+\alpha\mu_i}\right) ^{y_{i}}.\end{aligned}
\end{equation}The regression coefficients $\alpha$ and $\beta$ can be estimated using the maximum likelihood function:

\begin{equation}
\begin{aligned} L(\alpha,\beta) &= \prod_{i=1}^n p(y_i) \\
&= \prod_{i=1}^n \frac{\varGamma \left( y_{i}+\alpha^{-1} \right) }{\varGamma \left( y_{i}+\alpha^{-1} \right) }\left( \frac{1}{1+\alpha\mu_i}\right) ^{\alpha^{-1}}\left( \frac{ \alpha\mu_{i}}{1+\alpha\mu_i}\right) ^{y_{i}}.\end{aligned}
\end{equation}

\subsubsection{Deep Neural Network Model}
To predict the confirmed cases is extremely challenging since numerous known and unknown factors affect this pandemic spreading, such as traffic, population density, and how much people concern. However, accessing this information is not easy and may incur an additional cost. In this paper, we explore the most cutting-edge machine algorithms to predict the confirmed patients by Google trends data. Google trends data are able to indicate how much people concern about some specific topics, and provide an excellent opportunity to study the disease severity locally and globally. 
\begin{figure}[h]
    \centering
    \includegraphics[width=0.8\linewidth]{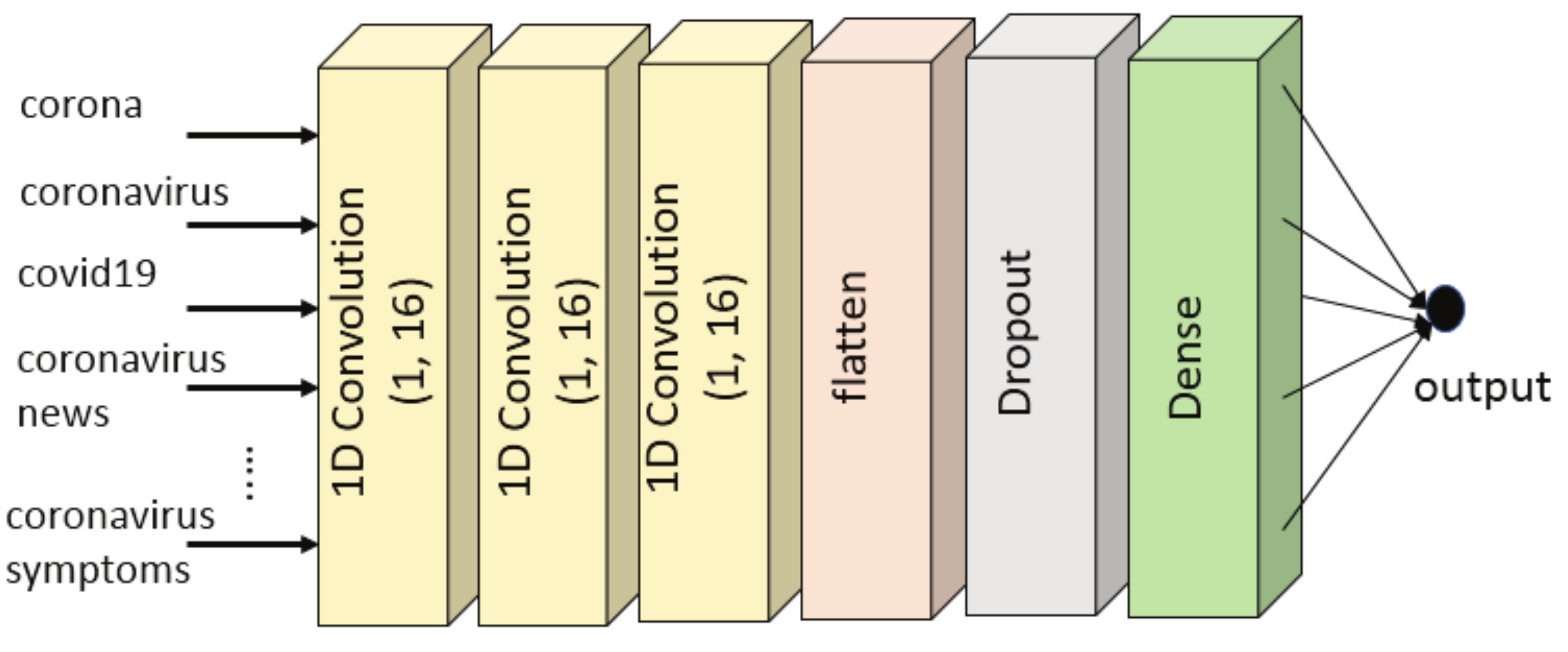}
    \caption{Deep neural network for confirmed cases prediction. }
    \label{fig:network}
\end{figure}

Since the pandemic depends on the skill of social distancing to prevent the spread, so there is limited influence from the temporal factor. Therefore, instead of using a temporal model, we decided to use one dimensional convolutional neural network as the core component in our prediction model. Figure \ref{fig:network} presents the overall architecture of the deep neural network prediction model. In particular, there are three connected 1D convolutional layers, a flatten layer, a drop out layer, and a dense layer. The first convolutional layer has a filter size of $16$, kernel size of $2$, strides of $3$, and a dilation rate of $1$. The second and third convolutional layer steps one step at a time with the same filter size of $16$, kernel size of $2$, while their dilation rates are $2$, and $4$, respectively. These layers have different dilation rates in order to help the network capture more contextual information in the feature map. Lastly, we appended the dropout layer of $5\%$ at the end before producing the output to avoid overfitting.

% \section{Evaluation}
% \subsection{Competing Approaches}
% \subsection{Comparison Metrics}
 
\section{Experiments and Results}
We start explaining the datasets. Next, we perform an analysis of Google search trends, related queries and the categories of the related topics with respect to new confirmed coronavirus cases. Finally, we show the results of forecasting the number of confirmed cases as there is a strong relationship between Google\'s trends (GTs) and the confirmed cases, that help to improve the performance of traditional forecasting algorithms as well as our proposed deep neural network.

\vspace{3mm}
\subsection{Datasets and Data Collection Procedure}
We crawl the Google Trend API to retrieve daily data from Jan 20, 2020 to March 23, 2020. The collection is done in both manner: a query for the data in a specific time range and a query for each day. Both of these types will produce trending of the search terms in a daily scale. However, the query with a time range will produce aggregated data for related queries and related topics. On the other hand, the daily query will generate daily information about different related queries and related topics. Hence, it helps to see the evolvement of such information through time and space. 
\begin{figure}[h]
    \centering
    \includegraphics[width=\linewidth]{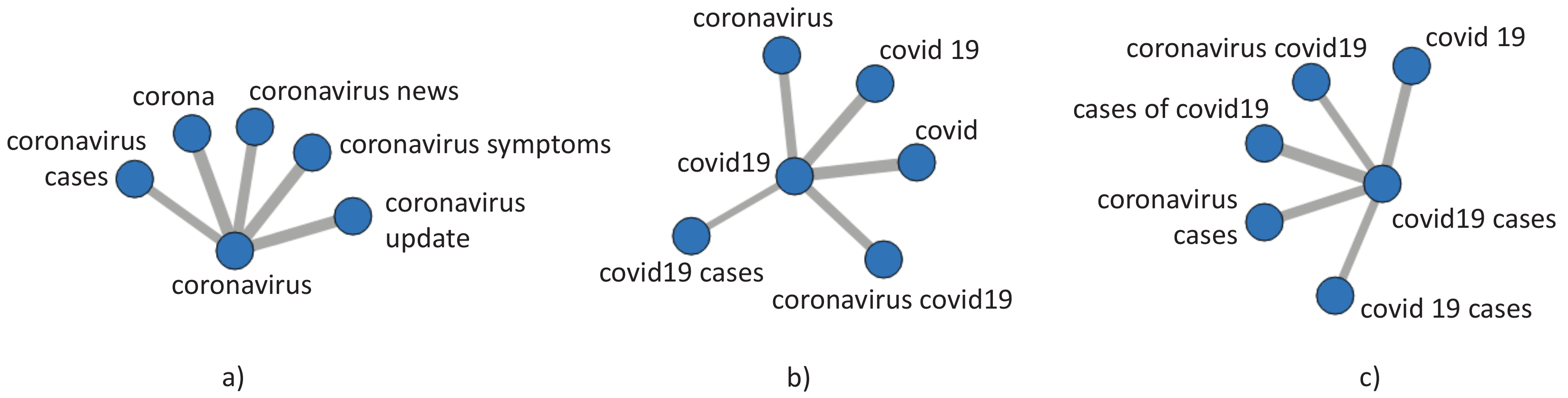}
    \caption{Selection of trending search queries and its expansion. The thickness of a connection represents the weight of the two terms.}
    \label{fig:trending-search-queries}
\end{figure}
Regarding the selection of search terms to derive its trending, the most obvious terms about coronavirus are used: \textit{coronavirus, covid19,} and \textit{covid19 cases}. In order to enrich the feature sets, we collect the related queries to the defined terms and extract the top five high-weighted related queries (Google Trends uses its algorithm to rank the related queries). These queries become the new search terms to pull its trending and other information from Google Trends. If the terms have appeared in other queries, it will be ignored. At the end, our dataset composes of the trending, related queries (top and rising related queries), related topics (top and rising related topics), and associated regions from $12$ trending search terms. Figure \ref{fig:trending-search-queries} demonstrates the three selected queries and its expansion to other related queries. The connection represents the comparison weight between the two terms based on search trend. For the forecasting algorithms, we randomly split the data with a ratio of $85\%$ for training and $15\%$. The deep neural network models use a dropout rate of $5\%$ to avoid overfitting. The same strategy is applied for country based prediction.

\vspace{5mm}
\subsection{Overview of Worldwide Corronavirus Cases}
\begin{figure}[t]
    \centering
    \includegraphics[width=0.95\linewidth]{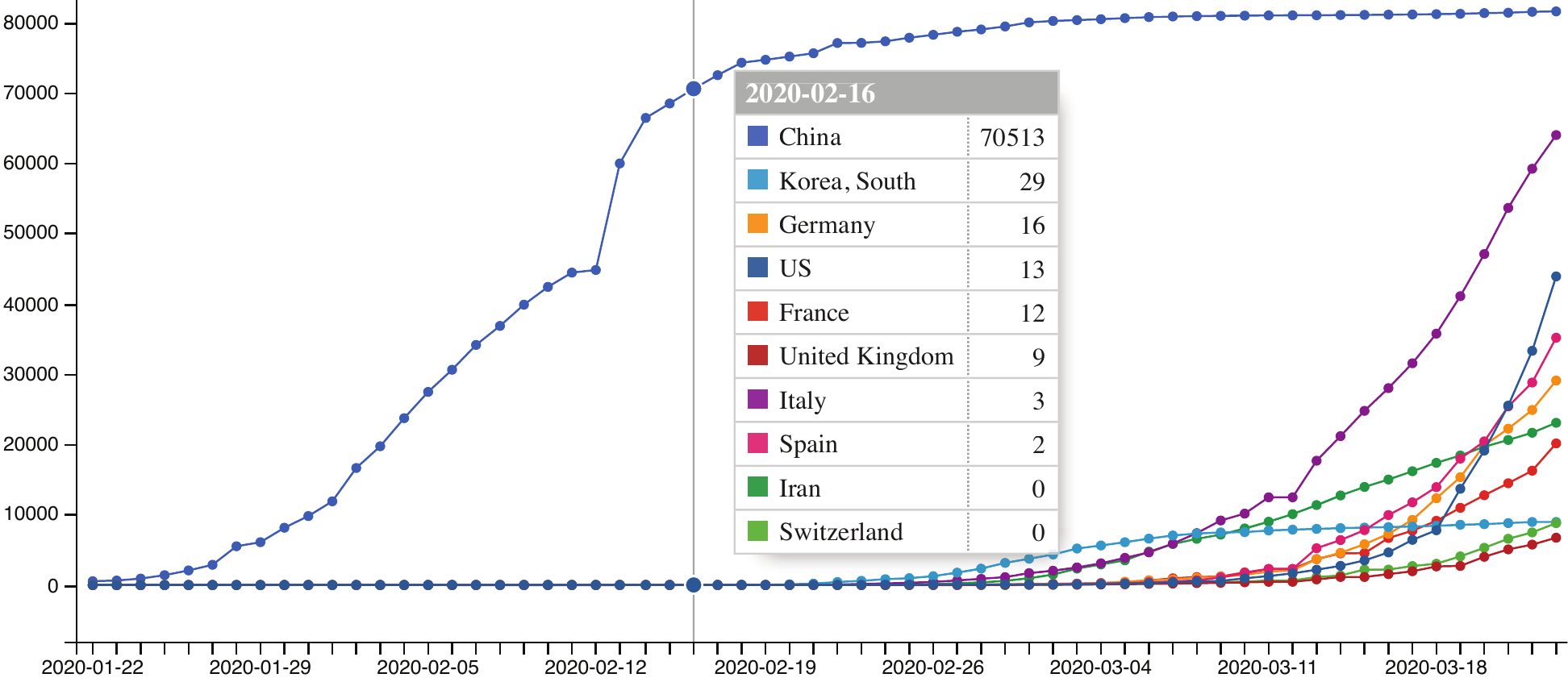}
    \caption{Cumulative confirmed cases of top 10 countries worldwide. 
    }
    \label{fig:top-10-confirmed-cases}
\end{figure}
Figure \ref{fig:top-10-confirmed-cases} presents the top $10$ countries having the most number of cumulative confirmed cases worldwide. The common trend of these countries is the exponential rise during the days of first confirmed cases. China has the earliest confirmed cases and keeps being the top in the number of confirmed cases until March 22, 2020. Italy, United States, and France cases are subsequent. 
When the number of confirmed cases in China is flattened by February 24, it was the start of the increment in Europe and the United States.
The confirmed cases have been dramatically spiked after 3/9. Aftermath, Italy hit the second in the world after China from that day on.
Starting the second week of March, the U.S. boosted exponentially in just two weeks to be the third largest confirmed cases.
Spain and Germany followed similar trajectories of the confirmed cases to rank the fourth and fifth, respectively. Iran was ranked the sixth even it started early than most of the countries.
South Korea and Switzerland ranked the same, though the slow start in Switzerland.
The least confirmed infected cases were in the UK as it was the last country of the top countries starting infected by COVID-19. During the period from the end of the fourth week of February until the third week of March, we can see that China has a different trajectory of confirmed cases than other counters. These countries began to record COVID-19 cases after China's new cases slowed down.
\vspace{2mm}
\subsection{Evolution of Internet Search Queries}
\begin{figure*}[h]
    \centering
    \includegraphics[width=0.95\linewidth, height=20cm]{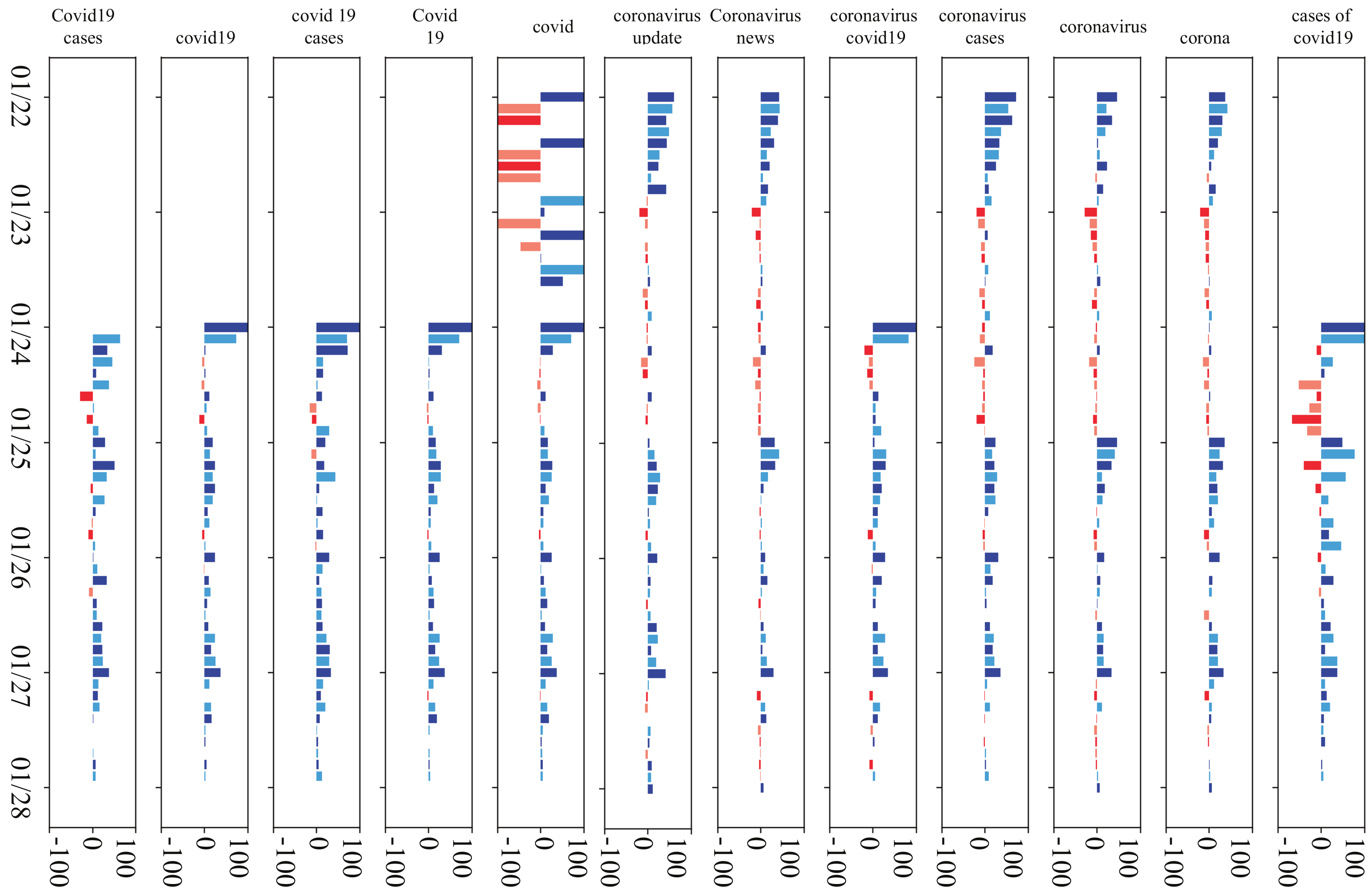}
    \caption{Daily interest change of each keyword. 
    }
    \label{fig:daily-change}
\end{figure*}

\begin{figure}[h]
    \centering
    \includegraphics[width=0.9\linewidth]{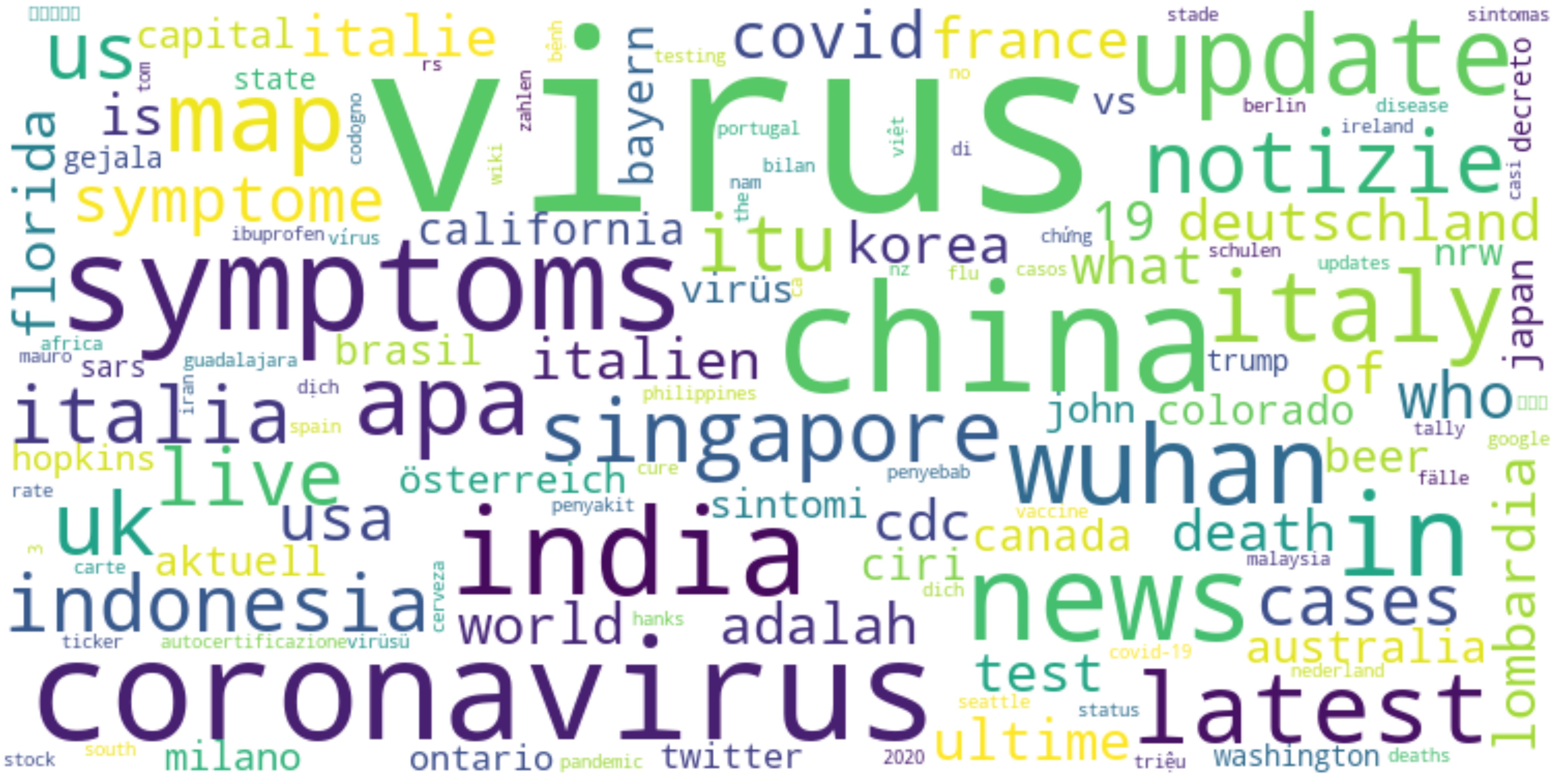}
    \caption{Wordcloud visualization of overall terms in the related queries.}
    \label{fig:general_related_queries_wordcloud}
\end{figure}

We studied the daily changes on the keywords as shown in Figure~\ref{fig:daily-change}. We use blue bars if the change is greater than zero; otherwise, reds. The darker colors are for even days and lighter colors are for odd days, to clarify the figure. COVID-19 was not known as coronavirus for many people during the start of the coronavirus outbreak. There are no trends for COVID-19 before 1/24/2020. Though WHO announced on 2/11/2020 the official name for the 2019 novel coronavirus to be COVID-19~\footnote{https://www.cdc.gov/coronavirus/2019-ncov/faq.html}, the trends used this official name after 14 days. It means that several people were not aware of this term before officially announced by the international health organization. 
We can see that COVID-19 terms are less fluctuation than corona terms.
The keyword "cases of covid19" started in the fourth week of January with high interest in the first couple of hours and dropped down most of that day. From the next day (1/25) until the end of the month, there was a trend of increasing usage of this keyword. Similar behaviors were for the keywords "coronavirus covid19" and "Covid19 cases". 
In general, the searched keywords in Google "Covid 19", "covid 19 cases", and "covid19" were increased exponentially during the last week of January.
Moreover, most of the keywords without "covid19" (i.e., "corona", "coronavirus", "coronavirus cases", "Coronavirus news", "Coronavirus update") started on the twenty second of January and decreased in the next two days. After that, they increased most of the time until the end of the month. Whereas, the keyword "covid" fluctuated on January twenty second until the mid of the next day. In general, it continued boosting until the end of this period. Figure~\ref{fig:general_related_queries_wordcloud} illustrates the word cloud of the related queries for all days in this study. It has several words in different sizes that represent the frequencies of the words. We can see that virus was the largest occurred word during this period. Also, the terms ("coronavirus", "china", "news", "update", "cases") were also frequent. These terms are parts of the related queries as shown in Figure~\ref{fig:daily-change}.
%%%%%%%%%%%%%%%%%%%%%%%%%%%%
\vspace{3mm}
\subsection{Evolution of Internet Search Related Queries}
\begin{figure}[t]
    \centering
    \includegraphics[width=0.9\linewidth]{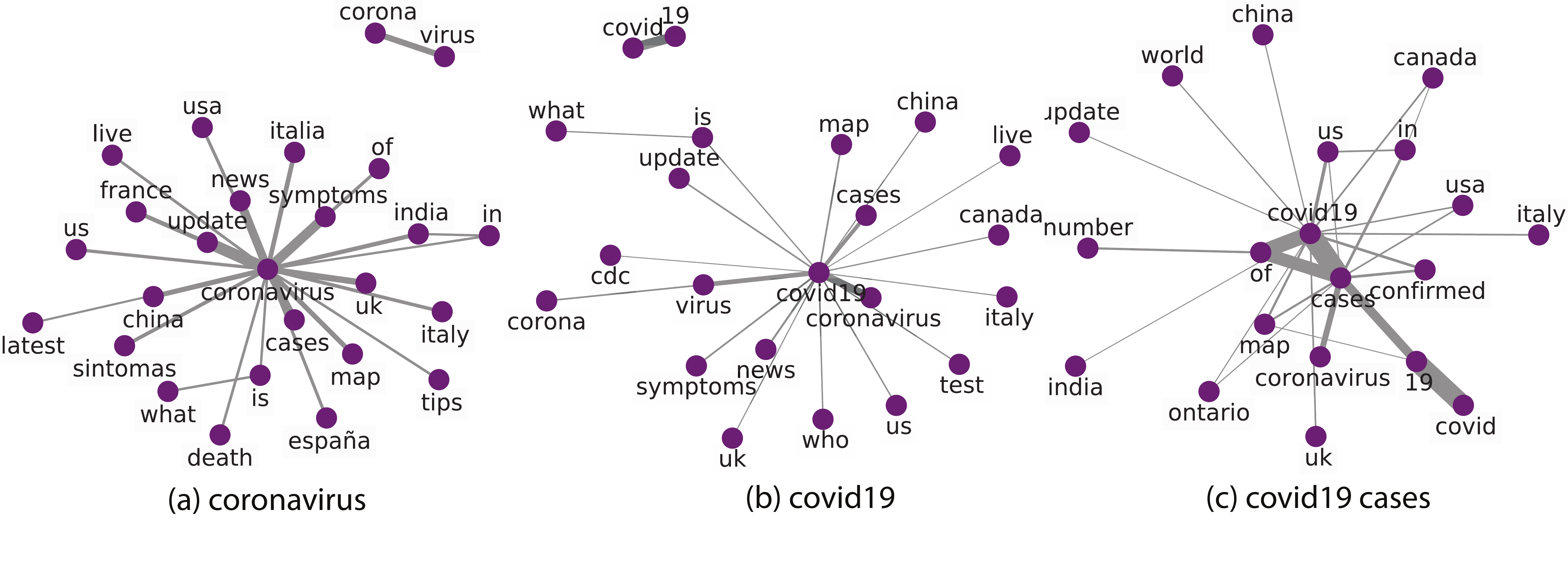}
    \caption{Word co-occurrence for the three keywords: (a) coronavirus, (b) covid19, and (c) covid19 cases.}
	  \label{fig:cooccurrence-queries}
\end{figure}

Word co-occurrences for the three keywords are shown in Figures~\ref{fig:cooccurrence-queries}. The keywords in the queries are: coronavirus, covid19, and covid19 cases. Coronavirus was related most with "cases", "uk", "symptoms", "news", and "update" as in Figure~\ref{fig:cooccurrence-queries}/a. People were concerned more with these terms for searching Google to understand and know more about coronavirus. The word co-occurrence for covid19 is shown in Figure~\ref{fig:cooccurrence-queries}/ b. It is clear that "cases", "virus", and "coronavirus" were the most correlated with this keyword. Also, the term "covid" was related to the number "19". Moreover, the keyword (covid19 cases) was more occurred in the terms of this composite keyword with its terms (covid19, cases) and the word "of" as shown in Figure~\ref{fig:cooccurrence-queries}/ c. The integer "19" was correlated with the term "cases".  

% \textcolor{blue}{Daily change of related queries with respect to a specific keyword}
% \textcolor{red}{Wordcloud and word co-occurrence [OK]}

%%%%%%%%%%%%%%%%%%%%%%%%%%%%
\vspace{2mm}
\subsection{Evolution of Internet Search Related Topics}
\begin{figure*}
    \centering
    \includegraphics[width=\linewidth]{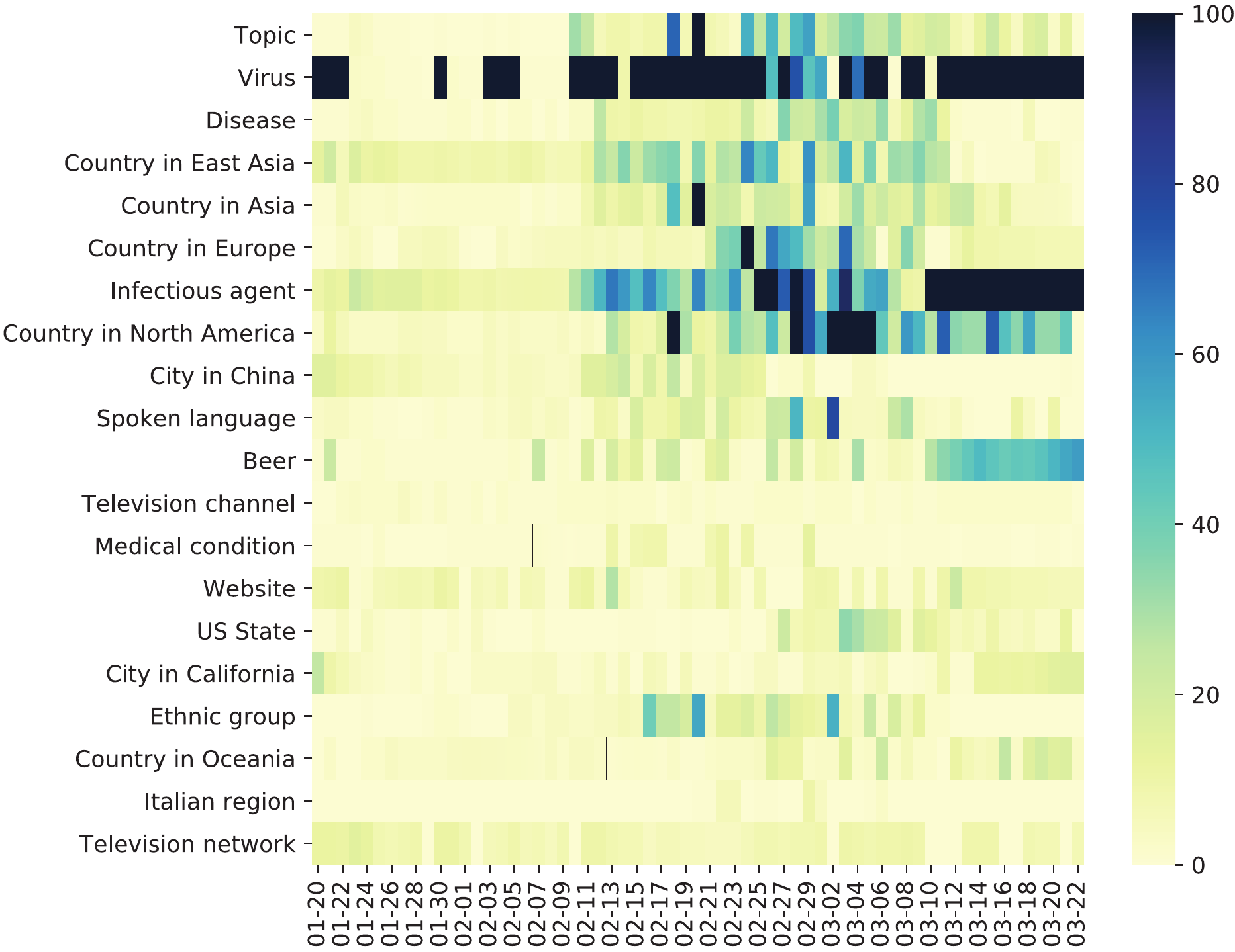}
    \caption{A heatmap visualization for the categories of related topics. The darker color, the higher mentioned frequencies.}
    \label{fig:top-related-topics}
\end{figure*}

Figure~\ref{fig:top-related-topics} shows the top 20 categories of topics over time between the second day of January until the twenty second of March. 
% Google has categorized the searched trends topics for These categories 
These entities (categories) are recognized (classified) by Google. In this paper, we used the related categories related to coronavirus or COVID-19.

People searched for the most popular category "Virus" on most days of this period. The second important searched category was "Infectious agent", especially in the mid of February until the third week of March. The interest of the category "Country in North America" increased the concern in March since more confirmed cases were discovered in the North American content.
% People were not interested in "Television channel" during this period.

The general category "Topic" named by Google was frequent at the beginning of the third week in February until the end of the period.
The category "Disease" occurred more at the end of February until the end of the second week of March.
Even though the Asia categories ("Country in East Asia" and "Country in Asia") were trends after February 11, the East Asia category was also frequent from the fourth week in January.
Categories related to cities in China were frequent from the start of the period of this study, until the mid of the ninth week of 2020 during the confirmed cases' curve in the country started to be flattened. Categories for places in California, Oceania, and Italy had fewer trends during most of the period of this research, while the category "US State" was the most frequent during the first week of March.

An article in CBS News on the first of March\footnote{https://www.cbsnews.com/news/cornavirus-corona-beer-they-have-nothing-to-do-with-each-other/} focused on a survey found that 38\% of American drinking beers would not order Corona Beer. 16\% of them thought that it might be a relation between drinking Corona Beer and COVID-19. This impacted the Google trends about beer. We can see that this trend was more frequent after 7 days of the article.

%%%%%%%%%%%%%%%%%%%%%%%%%%
\subsection{Confirmed Cases Prediction Performance}
We use root mean square error (RMSE) as the main metric to compare the performance of the prediction models. 
RMSE is a common metric to measure the differences between the actual value and the value predicted by the forecasting model.
Furthermore, other metrics like mean absolute error (MAE), mean absolute percentage error (MAPE), and r-square (R2) are also reported to relatively understand other aspects of the comparison.
MAE is a metric related to the average expected value of the loss (i.e., the loss of the absolute error).
MAPE is a regression metric used to measure the quality of a forecasting model, where the smaller is the better. In other words, it is the average accuracy ratio of the model.
R2 is an indication of how well the examples in a test dataset are possible to be predicted by a forecasting model.
The formulas of these evaluation metrics are: $RMSE = \sqrt{ \frac{1}{n} \sum_{i=1}^{n} (y_i - \hat{y_i})^2} $, $MAE = \frac{1}{n} \sum_{i=1}^{n} |y_i - \hat{y_i}|$, $MAPE = \frac{1}{n} \sum_{i=1}^{n} \frac{|y_i - \hat{y_i}|}{|y_i|}$,
$R2 = 1 - \frac{\sum_{i=1}^{n} |y_i - \hat{y_i}|}{\sum_{i=1}^{n} |y_i - \bar{y}|}$,
where $n$ is the size of the test dataset, $y_i$ is the actual number of confirmed cases, $\hat{y_i}$ is the predicted number of confirmed cases by a forecasting model based on the historical time series of trends and confirmed cases, and $\bar{y}$ represents the mean value $\forall{y_i}$, $i \in \{1, \dots, n\}$. 

\begin{table*}[htbp]
    \setlength{\tabcolsep}{20pt}
  \centering
  \caption{Confirmed cases forecast performance by learning algorithms.}
    \begin{tabular}{l|r|r|r|r}
    \toprule
          & \multicolumn{1}{c|}{RMSE} & \multicolumn{1}{c|}{MAE} & \multicolumn{1}{c|}{MAPE} & \multicolumn{1}{c}{R2} \\
    \midrule
    Linear & 1,685 & 847   & \textbf{0.39}  & 0.76 \\
    Linear + GT & 1,683 & 1,475 & 1.43  & 0.38 \\
    Negative Binomial & 1,645 & 1,194 & 1.74  & 0.77 \\
    Negative Binomial + GT & 1,494 & 1,373 & 1.11  & 0.51 \\
    Deep Neural Network & 1,595 & 1,317   & 0.48 & 0.76 \\
    Deep Neural Network + GT & \textbf{807} & \textbf{569} & 0.63  & \textbf{0.82} \\
    \bottomrule
    \end{tabular}%
  \label{tab:prediction-performance}%
\end{table*}%
\begin{figure}[t]
    \centering
    \includegraphics[width=0.95\linewidth]{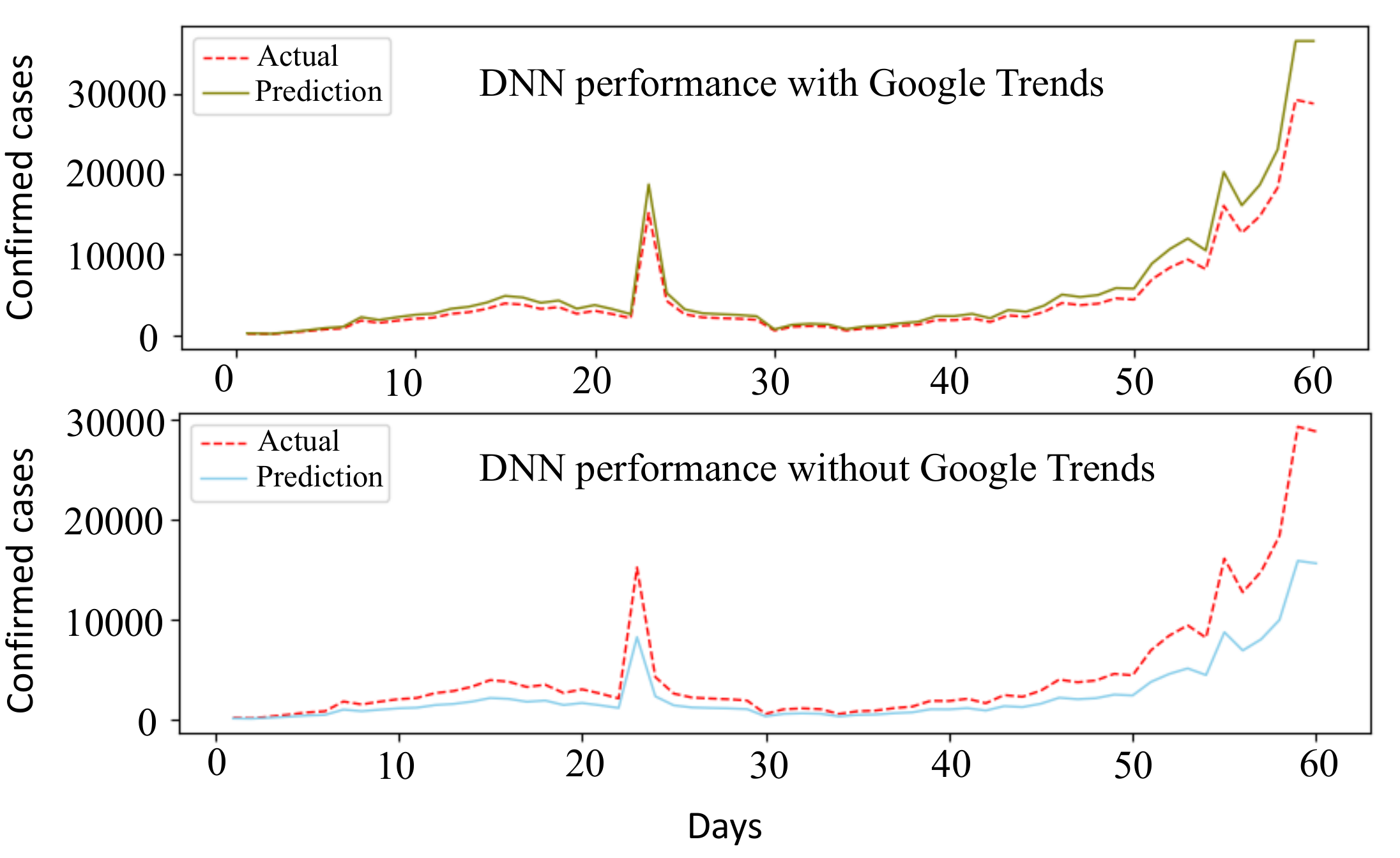}
    \caption{Actual versus prediction of confirmed cases using Deep Neural Network model. X axis represents the number of days since January 20, 2020.Y axis represents number of confirmed cases.
    }
    \label{fig:actual-vs-prediction}
\end{figure}
\textbf{World-wide prediction performance:}
Table \ref{tab:prediction-performance} presents the prediction performance of three types of learning algorithms (namely Linear, Negative Binomial, and Deep Neural Network) using Google trends and without using Google trends feature in the input (in this case, we use confirmed cases of previous day as the feature for the prediction model). The table shows that the addition of Google trends feature for Linear, Negative Binomial, and Deep Neural Network enhances themselves if not using Google trends feature. The linear model has a small improvement with RMSE increased from $1,685$ to $1,683$. Negative Binomial model presents better improvement of RMSE from $1,645$ to $1,494$. The Deep Neural Network model shows the best improvement with RMSE from $1,595$ to $807$, equivalent to $49\%$ enhancement. In all the variants, the Deep Neural Network model with Google trends outperforms other models.

Figure \ref{fig:actual-vs-prediction} demonstrates the visualization of new cases prediction compared with actual values using the Deep Neural Network model. We can see that the Deep Neural Network model is able to predict the pattern and show a smaller gap between the actual and the prediction lines compared with the version without using Google trends feature. Throughout the three models, we can see that Deep Neural Network using Google trends feature shows the best performance.

\begin{figure}[t]
    \centering
    \includegraphics[width=0.95\linewidth]{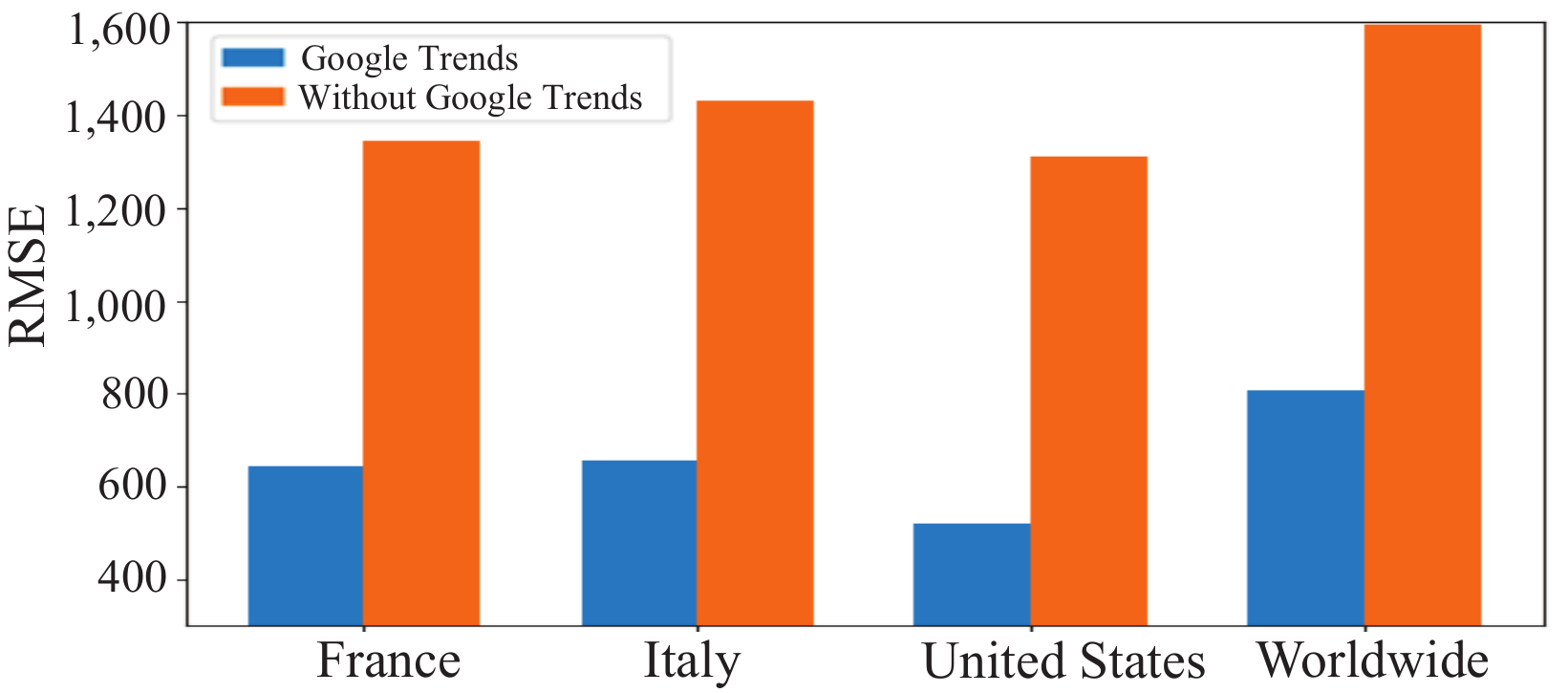}
    \caption{Deep neural network performance comparison across several countries such as France, Italy, United States, and Worldwide. We use one previous day as the feature when no google trends is used for prediction.
    }
    \label{fig:countries}
\end{figure}

\textbf{Country-based prediction performance:} We selected countries with increasing number of new confirmed coronavirus cases such as Italy, France, and United States to compare the model performance when using google trends and without google trends data. China is not reported in this case because all the queries do not correlate with the number of confirmed cases in China. It could due to the fact that users in China use its own search system to search the Internet instead of Google. Additionally, for the model that does not use google trends as features, we use confirmed cases of previous day to predict for the next day. Figure \ref{fig:countries} presents the performance of our proposed deep neural network model whether using or not using Google Trends as features.
From the Figure \ref{fig:countries}, we see that using Google Trends significantly enhances the deep neural network model. The RMSEs of the model using Google Trends in France, Italy, United States, and Worldwide are $643$, $657$, $520$, and $807$ while its performance when not using Google Trends are $1,346$, $1,432$, $1,310$, and $1,546$, respectively. We see that the model improves approximately two times in France, Italy, and Worldwide and about two and half times in United States datasets. The enhancement of the model performance across countries has confirmed that Google Trends can be a promising cheap source of information to predict new confirmed coronavirus cases.

\section{Conclusion}
In this paper, we present a spatio-temporal view on the relationship between Google search trends and the confirmed coronavirus cases. The framework supports visualization and analytic on the evolution of search trends, search queries, and related queries globally. Additionally, we explore the capability of Google search trends in predicting the number of confirmed cases for different types of learning models, namely Linear, Negative Binomial, and Deep Neural Network. The results show that Google search trends enhance the performance of the three different forecasting models, where the non-linear learning model, Deep Neural Network, has the best performance. Employing the Google search trends features fosters the performance of the Deep Learning approach more than 49\%. Thus, the potential data source is not only easy to access but also it is necessary to improve the performances of the employed forecasting models.

\bibliographystyle{frontiersinSCNS_ENG_HUMS} % for Science, Engineering and Humanities and Social Sciences articles, for Humanities and Social Sciences articles please include page numbers in the in-text citations
\bibliography{reference}

\end{document}